\newif\ifprocs
\procsfalse

\ifprocs
\documentclass{myllncs}
\else
\documentclass[11pt]{article}
\usepackage{amsthm}
\usepackage{fullpage}
\fi

\usepackage{amsmath,amssymb}
\usepackage{ifpdf}
\ifpdf    %
\usepackage{hyperref}
\else    %
\usepackage[hypertex]{hyperref}
\fi

\usepackage{color}
\usepackage{xspace}
\newcommand{\citep}[1]{\cite{{#1}}}
\newcommand{\citet}[1]{\cite{{#1}}}

\def\compactify{\itemsep=0pt \topsep=0pt \partopsep=0pt \parsep=0pt}

\ifprocs
\else
\newtheorem{theorem}{Theorem}[section]
\newtheorem{lemma}[theorem]{Lemma}
\newtheorem{corollary}[theorem]{Corollary}
\newenvironment{remark}{\noindent{\bf Remark.}\hspace*{1em}}{\bigskip}
\fi

\newcommand{\gam}{\gamma}
\newcommand{\nrm}[1]{\left\Vert #1 \right\Vert}
\newcommand{\sgn}{\operatorname{sgn}}

\newcommand{\E}{\mathbb{E}}
\renewcommand{\P}{\mathbb{P}}
\newcommand{\R}{\mathbb{R}}

\renewcommand{\SS}{\mathcal{S}}
\newcommand{\TT}{\mathcal{T}}

\newcommand{\FF}{\mathcal{F}}

\newcommand{\hN}{\hat{\mathcal{N}}}

\newcommand{\Dhat}{{d'}}
\newcommand{\D}{{d}}
\newcommand{\V}{\mathcal{V}}
\newcommand{\W}{\mathcal{W}}

\newcommand{\paren}[1]{\left( #1 \right)}
\newcommand{\sqprn}[1]{\left[ #1 \right]}

\newcommand{\set}[1]{\left\{ #1 \right\}}
\newcommand{\smset}[1]{\{ #1 \}}

\newcommand{\beq}{\begin{eqnarray*}}
\newcommand{\eeq}{\end{eqnarray*}}
\newcommand{\beqn}{\begin{eqnarray}}
\newcommand{\eeqn}{\end{eqnarray}}
\newcommand{\ben}{\begin{enumerate}}
\newcommand{\een}{\end{enumerate}}
\newcommand{\bit}{\begin{itemize}}
\newcommand{\eit}{\end{itemize}}
\newcommand{\hide}[1]{}
\newcommand{\oo}[1]{\frac{1}{#1}}
\newcommand{\eps}{\epsilon}

\newcommand{\ceil}[1]{\ensuremath{\left\lceil#1\right\rceil}}

\newcommand{\distor}{\eta}
\newcommand{\rade}[2]{\hat R_n(#1;\set{#2_i})}

\usepackage{bbm}
\newcommand{\chr}{\boldsymbol{\mathbbm{1}}} %
\newcommand{\pred}[1]{\chr_{\left\{ #1 \right\}}}

\newcommand{\setpm}{\set{-1,1}}
\newcommand{\prl}{^\parallel}
\newcommand{\prp}{^\perp}

\newcommand{\X}{\mathcal{X}}
\newcommand{\calZ}{\mathcal{Z}}
\newcommand{\calN}{\mathcal{N}}

\newcommand{\F}{{F}}

\renewcommand{\H}{H}

\DeclareMathOperator{\ddim}{ddim}
\DeclareMathOperator{\diam}{diam}
\def\eps{\varepsilon}

\newcommand{\aset}[1]{\{{#1}\}}

\newcommand{\abs}[1]{\left| #1 \right|}

\newcommand{\bepf}{\begin{proof}}
\ifprocs
\newcommand{\enpf}{\qed\end{proof}}
\else
\newcommand{\enpf}{\end{proof}}
\fi

\title{Adaptive Metric Dimensionality Reduction}
\author{
Lee-Ad Gottlieb%
\thanks{Ariel University, Ariel, Israel.
Email: \texttt{leead@ariel.ac.il}.
}
\and  
Aryeh Kontorovich%
\thanks{Ben-Gurion University of the Negev, Beer Sheva, Israel.
Email: \texttt{karyeh@cs.bgu.ac.il}.
This research was partially supported by the Israel Science Foundation 
(grant \#1141/12) and
a Yahoo Faculty award.
}
\and 
Robert Krauthgamer%
\thanks{Weizmann Institute of Science, Rehovot, Israel.
Email: \texttt{robert.krauthgamer@weizmann.ac.il}.
This work was supported in part by a US-Israel BSF grant \#2010418, 
an Israel Science Foundation grant \#897/13,
and by the Citi Foundation.}
}

\begin{document}

\maketitle

\begin{abstract}
We study adaptive data-dependent dimensionality reduction 
in the context of supervised learning in general metric spaces.
Our main statistical contribution is a generalization bound 
for Lipschitz functions in metric spaces that are doubling, or nearly doubling.
On the algorithmic front, we describe an analogue of PCA for metric spaces:
namely an efficient procedure that approximates the data's intrinsic dimension,
which is often much lower than the ambient dimension.
Our approach thus leverages the dual benefits of low dimensionality: 
(1) more efficient algorithms, e.g., for proximity search, and 
(2) more optimistic generalization bounds.
\end{abstract}

\section{Introduction}

Linear classifiers play a central role in supervised learning, 
with a rich and elegant theory.
This setting assumes
data is represented as points in a Hilbert space,
either explicitly as feature vectors or implicitly via a kernel.
A significant strength of the Hilbert-space model is its inner-product structure,
which has been exploited statistically and algorithmically
by sophisticated techniques from geometric and functional analysis,
placing the celebrated hyperplane methods on a solid foundation.
However, the success of the Hilbert-space model 
obscures its limitations
---
perhaps the most significant of which
is that it cannot 
represent many norms and distance functions that arise naturally in applications.
Formally, metrics such as $L_1$, earthmover, and edit distance 
cannot be embedded into a Hilbert space without distorting distances
by a large factor \cite{Enflo69,NS07,AK10}.
Indeed, the last decade has seen a growing interest and success
in extending the theory of linear classifiers
to Banach spaces and even to general metric spaces,
see e.g. \citep{DBLP:conf/colt/MicchelliP04,DBLP:journals/jmlr/LuxburgB04,DBLP:journals/jcss/HeinBS05,der-lee-banach,ZXZ09}.

A key factor in the performance of learning is the dimensionality of the data,
which is known to control the learner's efficiency, 
both statistically, i.e.\ sample complexity \cite{shwartz2014understanding},
and algorithmically, i.e.\ computational runtime \cite{KL04}.
This dependence on dimension is true not only for Hilbertian spaces,
but also for general metric spaces,
where both the sample complexity and the algorithmic runtime can be bounded
in terms of the covering number or the doubling dimension 
\citep{DBLP:journals/jmlr/LuxburgB04,DBLP:journals/tit/GottliebKK14,semimetric15}.

In this paper, we demonstrate that
the learner's statistical and algorithmic
efficiency can be controlled by
the data's \emph{intrinsic dimensionality}, 
rather than its \emph{ambient dimension} (e.g., the representation dimension).
This provides rigorous confirmation for the informal insight that real-life data
(e.g., visual or acoustic signals) can often be learned efficiently
because it tends to lie close to low-dimensional manifolds, even when represented
in a high-dimensional feature space.
Our simple and general framework quantifies 
what it means for data to be \emph{approximately} low-dimensional,
and shows how to leverage this for computational and statistical gain.

Most generalization bounds depend on the intrinsic dimension, 
rather than the ambient one,
when the training sample lies \emph{exactly} on a low-dimensional subspace.
This phenomenon is indeed immediate in generalization bounds obtained via
the empirical Rademacher complexity 
\citep{DBLP:journals/jmlr/BartlettM02,MR1892654},
but we are not aware of rigorous analysis that specifies such bounds to 
the case where the sample is ``close'' to a low-dimensional subspace.

\paragraph{Our contribution.}
We present learning algorithms and generalization bounds that adapt 
to the {\em intrinsic} dimensionality of the data,
and can exploit a training set that is \emph{close} to being low-dimensional
for improved runtime complexity and statistical accuracy.

We start with the scenario of a Hilbertian space, which is technically simpler.
Let the observed sample be $(x_1,y_1),\ldots,(x_n,y_n)\in \R^N\times\setpm$,
and suppose that $\aset{x_1,\ldots,x_n}$ 
is close to a low-dimensional linear subspace $T\subset \R^N$,
in the sense that its distortion 
$\distor = \frac1n \sum_i \nrm{x_i-P_T(x_i)}_2^2$ is small,
where $P_T:\R^N\to T$ denotes orthogonal projection onto $T$.
We prove in Section \ref{sec:warmup}
that when $\dim(T)$ and the distortion $\distor$ are small,
a linear classifier generalizes well 
regardless of the ambient dimension $N$ or the separation margin.
Implicit in our result is a statistical tradeoff between 
the reduced dimension and the distortion,
which can be optimized (for the sample at-hand) by performing PCA;
see Corollary \ref{cor:euclid-hinge}.

Our approach quantifies the statistical effect of using PCA,
which is commonly done to denoise or improve algorithmic runtime,
prior to constructing a linear classifier. 
We show that for low intrinsic dimension
and when the PCA cutoff value is chosen properly,
a linear classifier trained on points produced by PCA incurs only a 
small loss in generalization bounds; see Equation 
\eqref{eq:euclid-hinge-proj-sample}.%
\footnote{However, our bounds do not explain some reports of
empirical {\em improvements} achieved by PCA preprocessing.}

\medskip
We then develop this approach significantly beyond the Euclidean case,
to the much richer setting of general metric spaces.
A completely new challenge that arises here is the algorithmic part,
because no metric analogue to dimension reduction via PCA is known.
Let the observed sample be $(x_1,y_1),\ldots,(x_n,y_n)\in \X\times\setpm$,
where $(\X,\rho)$ is some metric space.
The statistical framework proposed by von Luxburg and Bousquet
\cite{DBLP:journals/jmlr/LuxburgB04},
where classifiers are realized by Lipschitz functions,
was extended by Gottlieb et al.\
\cite{DBLP:journals/tit/GottliebKK14} to obtain 
generalization bounds and algorithmic runtime 
that depend on the metric's doubling dimension, denoted $\ddim(\X)$ 
(see Section \ref{sec:prelims} for definitions).
The present work makes a considerably less restrictive assumption
--- that the sample points lie {\em close} to some low-dimensional set.

We establish in Section \ref{sec:general} new generalization bounds
for the scenario where there exists some multiset 
$\tilde S=\set{\tilde x_1,\ldots,\tilde x_n}$ 
of low doubling dimension, whose distortion 
$\distor = \oo n\sum_i \rho(x_i,\tilde x_i)$ is small.
In this case, the Lipschitz-extension classifier will generalize well, 
regardless of the ambient dimension $\ddim(\X)$;
see Theorem \ref{thm:rade-lip-margin}.
These generalization bounds feature a tradeoff between the 
intrinsic dimension and the distortion, 
which is separate from the tradeoff between the classifier's empirical loss
and its smoothness (Lipschitz constant).
Here, the first tradeoff must be optimized before addressing the second one.

Hence, we address in Section \ref{sec:alg} the computational problem 
of finding (in polynomial time) a near-optimal point set $\tilde S$,
given a bound on $\eta$.
Formally, we devise an algorithm that achieves a bicriteria approximation,
meaning that $\ddim(\tilde S)$ and $\distor$ of the reported solution
exceed the values of a target low-dimensional solution
by at most a constant factor; 
see Theorem \ref{thm:bicriteria}.
Using this algorithm, one can optimize the dimension-distortion tradeoff 
(for the sample at-hand) within a constant factor.
Having determined these values, one can compute a Lipschitz classifier 
that optimizes the tradeoff between empirical loss and smoothness 
(in the generalization bound), 
by running the algorithm of Gottlieb et al.\
\citet{DBLP:journals/tit/GottliebKK14} on the sample.
One may instead run this algorithm
on the modified training set (low-dimensional $\tilde S$), and then 
the runtime of constructing the classifier and evaluating it on new points 
will depend on the intrinsic dimension instead of the ambient dimension. 
We show that this incurs only a
small loss in the generalization bound; see Equation 
\eqref{eq:rade-lip-margin-perturbed-sample}.

\paragraph{Related Work.}

The topic of dimensionality reduction, even restricted to learning theory,
is far too vast to adequately survey within the scope of this paper
(for recent surveys, see 
e.g.\ \cite{nonlin-dimred-book,DBLP:journals/ftml/Burges10}).
We are only aware of dimensionality results with provable guarantees
for the Euclidean case
--- mainly in the context of improving algorithmic runtimes ---
achieved by projecting data
onto a random low-dimensional subspace,
see e.g.\ \citep{DBLP:journals/ml/BalcanBV06, DBLP:conf/nips/RahimiR07,
DBLP:conf/icml/ShiSHH12,
DBLP:conf/aistats/PaulBMD13}.
On the other hand, data-dependent dimensionality reduction techniques
have been observed empirically to improve and speed up classification performance.
For instance, PCA may be applied as a preprocessing step before 
learning algorithms such as SVM,
or the two can be put together into a combined algorithm,
see e.g.\ 
\citep{DBLP:journals/jmlr/BiBEBS03, FBM04, HA07, DBLP:journals/tsp/VarshneyW11}.

There is little existing rigorous work on dimension reduction in general metric spaces.
One heuristic approach is
Multi-Dimensional Scaling (MDS), 
which may be viewed as a generalization of PCA \cite{borg05}.
Given a finite metric space $\X$ and a target dimension $d$, 
MDS attempts to find a low-distortion
embedding of $\X$ into $\R^d$.
Thus, its usefulness is limited to
metrics that are ``nearly'' Euclidean,
and it will not be effective in general on
inherently non-Eucliean metrics such as $L_1$,
earthmover, and edit distance.
Furthermore, MDS optimizes a highly non-convex function, and
we are not aware of any performance guarantees for this algorithm ---
even for input metrics that are nearly Euclidean.
A different metric dimension reduction problem was considered in
\cite{DBLP:conf/approx/GottliebK10}:
{\em removing} from an input set $S$ as few points as possible,
with the goal of retaining
a large subset of low doubling dimension.
While close in spirit, their objective is technically different from ours,
and the problem seems to require rather different techniques.

Algorithmic questions aside, there is the issue {\em statistical} benefits 
gained by reducing the dimensionality --- or merely by realizing that the data
is effectively low-dimensional, or nearly so.
Here again,
previous work has mainly addressed statistical efficiency in Hilbertian spaces.
For SVM classification,
a large body of work derives generalization bounds from the empirical properties
of the kernel integral operator --- its spectrum, entropy numbers of its range on the unit ball, and 
other measures of ``effective dimension''
\cite{SSSW99,
DBLP:journals/tit/WilliamsonSS01,
DBLP:journals/jmlr/Mendelson03,
blanchard2008,
DBLP:journals/focm/HsuK014}. 
Recently, various 
geometric notions
of ``low intrinsic dimensionality'' were proposed by Sabato et al.\
\cite{DBLP:conf/nips/SabatoST10}
for the purpose of providing improved bounds on the sample complexity of large-margin learning.
In addition to ``passively'' benefitting from low intrinsic dimensionality,
one could actively seek to reduce the data-dimension,
and some attempts to quantify the resulting statistical benefits
have been made
\cite{MR2450775,DBLP:conf/icml/MosciRV07}.
In the context of regression, \cite{bickel07}
and the series of works cited therein observe that
``identifying
intrinsic low dimensional structure from a seemingly high dimensional source''
yields more optimistic minimax rates.

In general metric spaces, Dasgupta, Kpotufe and coauthors have observed in a series of works
that various $k$-NN methods automatically adapt to low data dimensionality 
\cite{NIPS2009_1009,
DBLP:conf/nips/Kpotufe11,
KpotufeDasgupta2012,
DBLP:conf/nips/KpotufeG13}. The analysis assumes that the data lies exactly (or almost so)
on a low-dimensional subset, 
and is not amenable to the sort of dimension-distortion tradeoff we study in this paper.
Furthermore, the classifier we employ here is based on Lipschitz-extension,
which eventually boils down to $1$-NN, 
and enjoys several advantages over $k$-NN: 
the effect of approximate proximity search on generalization performance may be quantified
\cite{DBLP:journals/tit/GottliebKK14} 
and near-optimal sample compression may be achieved \cite{DBLP:journals/corr/GottliebK14},
without sacrificing the Bayes consistency of $k$-NN \cite{DBLP:journals/corr/KontorovichW14a}.

\section{Definitions and notation}
\label{sec:prelims}
We use standard notation and definitions throughout,
and assume a familiarity with the basic notions of Euclidean and normed spaces.
We write $\pred{\cdot}$ 
for the indicator function of the relevant predicate
and
$\sgn(x):=2\cdot\pred{x\ge0}-1$.

\paragraph{Metric spaces.}
A {\em metric} $\rho$ on a set $\X$ is a positive symmetric function
satisfying the triangle inequality $\rho(x,y)\leq \rho(x,z)+\rho(z,y)$; together the two comprise the metric space $(\X,\rho)$.
A
function $f:\X\to\R$
is said to be $L$-Lipschitz if
$\abs{f(x)-f(y)}\leq L\cdot\rho(x,y)$ for all $x,y\in\X$.
The {\em diameter}, denoted by $\diam(\X)$, is defined to be $\sup_{x,x'\in\X}\rho(x,x')$.

\paragraph{Doubling dimension.}
For a metric $(\X,\rho)$, let $\lambda_\X>0$
be the smallest value such that every
ball in $\X$ can be covered by $\lambda_\X$ balls of half the radius.
Then $\lambda_\X$ is the {\em doubling constant} of $\X$, and
the {\em doubling dimension} of $\X$ is defined as $\ddim(\X):=\log_2(\lambda_\X)$.
It is well-known that while a $d$-dimensional Euclidean space,
or any subset of it, has doubling dimension $O(d)$;
however, low doubling dimension is strictly more general 
than low Euclidean dimension, see e.g.\ \cite{DBLP:conf/focs/GuptaKL03}.

We will use $\abs{\cdot}$ to denote the cardinality of finite metric spaces.

\paragraph{Covering numbers.}
The $\eps$-covering number of a metric space $(\X,\rho)$, denoted $\calN(\eps,\X,\rho)$, is defined as the smallest number 
of balls
%\footnote{
%As pointed out by a referee, these 
%%are ``centerless'' balls that 
%balls
%need not be entirely contained in $\X$.
%} 
of radius $\eps$ that suffices to cover $\X$.
The balls may be centered at points of $\X$, or of points in an ambient space including $\X$.
Covering numbers may be estimated
by repeatedly invoking the doubling property (see e.g. \cite{KL04}):
\begin
{lemma}
\label{lem:doublpack}
If $(\X,\rho)$ is a 
metric space
with $\ddim(\X)<\infty$ and $\diam(\X)<\infty$,
then for all $0<\eps<\diam(\X)$,
\beq
\calN(\eps,\X,\rho)
&\leq& \paren{\tfrac{2\diam(\X)}{\eps}}
^{\ddim(\X)}.
\eeq
\end{lemma}

\paragraph{Learning.}
Our setting in this paper is the {\em agnostic PAC} learning model,
see e.g.\ \citet{mohri-book2012}, 
where labeled examples $(X_i,Y_i)$ are drawn independently from
$\X\times\set{-1,1}$ according to some unknown probability distribution $\P$.
The learner, having observed $n$ 
labeled examples,
produces a hypothesis
$h:\X\to\set{-1,1}$. The {\em generalization error}
$\P(h(X)\neq Y)$
is the probability of misclassifying a new point,
although sometimes the 0-1 penalty here is replaced with a loss function.
Most 
generalization
bounds 
consist of
a {\em sample error} term 
(approximately corresponding to {\em bias} in statistics), which is the fraction 
of observed examples misclassified by $h$ and a 
{\em hypothesis complexity} term (a rough analogue of {\em variance} in statistics) 
which measures the richness of the class of all admissible hypotheses \citep{MR2172729}.
A data-driven procedure for selecting the correct hypothesis complexity is known as {\em model selection}
and is typically performed by some variant of Structural Risk Minimization 
\citep{DBLP:journals/tit/Shawe-TaylorBWA98}
--- an analogue of the bias-variance tradeoff in statistics.
The measure-theoretic technicalities associated with taking suprema over uncountable function classes are typically
glossed over in learning literature. However, we note in passing that 
if $\X$ is a compact metric space and
$\FF\subset\R^\X$ is a collection of Lipschitz functions,
then $\FF$ contains a countable $\FF'\subset\FF$ such that every member of $\FF$ is a pointwise
limit of a sequence in $\FF'$, which more than suffices to guarantee the measurability of the supremum
\cite{dudley1999uniform}.

\paragraph{Rademacher complexity.}
For any $n$ points $z_1,\ldots,z_n$ in 
some set
$\calZ$ and any collection of functions $\F$ mapping $\calZ$ to a bounded range,
define the Rademacher complexity of $\F$ evaluated at the $n$ points as
$$
\rade{\F}{z}
= 
\oo n \E\sup_{g\in\F}
\sum_{i=1}^n \sigma_i g(z_i),
$$
where 
$\sigma_i$ 
are 
iid random variables 
that take on $\pm1$ with probability $1/2$.
The seminal work of \cite{DBLP:journals/jmlr/BartlettM02} and \cite{MR1892654} 
established the central role
of Rademacher complexities in generalization bounds. 

The Rademacher complexity of a binary function class 
$\F$ may be controlled by the VC-dimension $d$ of $\F$ through an application of Massart's and Sauer's lemmas:
\beqn
\label{eq:rade-vc}
\rade{\F}{z}
\le \sqrt{\frac{2d\log(en/d)}{n}}.
\eeqn
Considerably more delicate bounds
may be obtained by estimating the covering numbers and using Dudley's chaining integral:
\beqn
\label{eq:dudley-int}
\rade{\F}{z} \le
\inf_{\alpha\ge0}\paren{4\alpha+12\int_\alpha^\infty\sqrt{\frac{\log \calN(t,\F,\nrm{\cdot}_2)}{n}}dt}.
\eeqn
A proof of (\ref{eq:rade-vc}) may be found in \cite{mohri-book2012}, and (\ref{eq:dudley-int})
is (essentially) contained in \cite{kakade-tewari-lecture}.

\section{Adaptive dimensionality reduction: Euclidean case}
\label{sec:warmup}
In this section, we illustrate our statistical approach in the familiar setting of Euclidean spaces, 
where the algorithmic problem of finding a low-distortion subspace is solved by PCA.
Consider the problem of supervised classification in $\R^N$
by linear hyperplanes,
where $N\gg1$.
The training sample is
$(X_i,Y_i)$, $i=1,\ldots,n$, with $(X_i,Y_i)\in\R^N\times\set{-1,1}$.
Since this set is finite, there is no loss of generality in normalizing
$\nrm{X_i}_2\le1$.
We consider the hypothesis class 
$\H=\set{x\mapsto \sgn(w\cdot x) : \nrm{w}_2\le1}$.
Absent
additional assumptions on the data, this is a high-dimensional learning problem
with a costly sample complexity. 
Indeed, the VC-dimension of linear hyperplanes in $N$ dimensions is $N$.
If, however, it turns out that the data actually lies on a 
$k$-dimensional subspace 
of $\R^N$,
Eq.\ (\ref{eq:rade-vc}) 
implies
that $\rade{\H}{X}\le \sqrt{{2k\log(en/k)}/{n}}$,
and hence a much better generalization for $k\ll N$.
A more common distributional assumption is that of large-margin separability.
In fact, the main insight articulated in
\citet{DBLP:conf/slsfs/Blum05}
is that data separable by margin $\gamma$ effectively lies in an $\tilde O(1/\gamma^2)$-dimensional space.

Suppose now that the data lies ``close'' to
a low-dimensional subspace.
Formally, we say that the data $\set{X_i}$ is
\emph{$\distor$-close} to a subspace $T\subset\R^N$
if
$
\frac1n \sum_{i=1}^n \nrm{X_i - P_T(X_i)}_2^2 \le \distor
$
(recall that $P_T(\cdot)$ denotes the orthogonal projection onto the subspace $T$).
Whenever this
holds,
the Rademacher complexity can be bounded in terms of $\dim(T)$ and $\distor$ alone (Theorem \ref{thm:euclidean}).
As a
consequence,
we obtain a bound on the expected hinge-loss
(Corollary \ref{cor:euclid-hinge}).

\begin{theorem}
\label{thm:euclidean}
Let $X_1,\ldots,X_n$ lie in $\R^N$ with $\nrm{X_i}_2\le1$
and define the
function class
$\F=\set{x\mapsto w\cdot x: \nrm{w}_2\le1}$.
Suppose that 
the data $\set{X_i}$ is $\distor$-close to
some subspace $T\subset\R^N$ and $\distor \ge 0$.
Then
$
\rade{\F}{X}
\le 17\sqrt\frac{\dim(T)}{n} + \sqrt\frac{\distor}{n}.
$
\end{theorem}
\begin{remark}
Notice that our Rademacher complexity bound is independent of
the ambient dimension $N$.
Our bound exhibits a tension between $\dim(T)$ and $\distor$,
--- as we seek a lower-dimensional approximation,
we are liable to incur a larger distortion ---
and optimizing the tradeoff between them amounts to choosing a PCA cutoff value.
\end{remark}

\bepf
Denote by $S\prl=(X_1\prl,\ldots,X_n\prl)$ and $S\prp=(X_1\prp,\ldots,X_n\prp)$ the parallel
and perpendicular components of the points $\set{X_i}$ with respect to $T$. 
Note that each $X_i$ has the unique
decomposition $X_i=X_i\prl+X_i\prp$.
We first decompose the Rademacher complexity into ``parallel'' and ``perpendicular'' terms:
\beqn
\rade{\F}{X}
&=& \oo n \E_\sigma\sqprn{ \sup_{\nrm{w}\le1} \sum_{i=1}^n \sigma_i(w\cdot X_i)} \nonumber\\
&=& \oo n \E_\sigma\sqprn{ \sup_{\nrm{w}\le1} w\cdot\sum_{i=1}^n  \sigma_i(X_i\prl+X_i\prp)} \nonumber\\
&\le& \hat R_{n}(\F;S\prl)+\hat R_{n}(\F;S\prp).
\label{eq:prl-perp-decomp}
\eeqn
We then proceed to bound the two terms
in (\ref{eq:prl-perp-decomp}). To bound the first term,
note that $\F$ restricted to $T$ is a 
function class with linear-algebraic dimension $\dim(T)$, and furthermore our assumption that the data
lies in the unit ball implies that the range of $F$ is bounded by $1$ in absolute value. Hence, the classic
covering number estimate (see \citet{MR1965359})
\beqn
\label{eq:l2-cov-3tk}
\calN(\F,t,\nrm{\cdot}_2) \le \paren{\frac{3}{t}}^{\dim(T)},
\qquad
\forall t\in (0,1)
\eeqn
applies. Substituting (\ref{eq:l2-cov-3tk}) into Dudley's integral (\ref{eq:dudley-int}) yields
\beqn
\hat R_{n}(\F;S\prl) &\le& 12\int_0^\infty\sqrt{\frac{\log \calN(t,\F,\nrm{\cdot}_2)}{n}}dt \nonumber\\
&\le& 12\int_0^1 \sqrt{\frac{
\dim(T)\log(3/t)
}{n}}dt \le 17\sqrt\frac{\dim(T)}{n}.
\label{eq:euclid-dudley}
\eeqn

The second term in (\ref{eq:prl-perp-decomp}) is bounded via a standard calculation:
\beq
\hat R_{n}(\F;S\prp) &=& \oo n \E_\sigma\sqprn{ \sup_{\nrm{w}_2\le1} w\cdot\sum_{i=1}^n  \sigma_iX_i\prp} 
=
\oo n \E_\sigma \nrm{\sum_{i=1}^n  \sigma_iX_i\prp}_2 \\
&\le& \oo n \paren{\E_\sigma \nrm{\sum_{i=1}^n  \sigma_iX_i\prp}_2^2}^{1/2},
\eeq
where the second equality follows from the dual characterization of the $\ell_2$ norm
and the inequality is Jensen's.
Now by independence of the Rademacher variables $\sigma_i$,
\beq
\E_\sigma \nrm{\sum_{i=1}^n  \sigma_iX_i\prp}_2^2
&=&
\E_\sigma \sum_{1\le i,j\le n} \sigma_i\sigma_j(X_i\prp\cdot X_j\prp)
=
\sum_{i=1}^n \nrm{X_i\prp}_2^2\\
&=&
\sum_{i=1}^n \nrm{P_T(X_i)-X_i}_2^2 \le n\distor,
\eeq
which implies
$\hat R_{n}(\F;S\prp) 
\le \sqrt{{\distor}/{n}}$
and together with (\ref{eq:prl-perp-decomp}) and (\ref{eq:euclid-dudley}) proves the claim.
\enpf

\begin{corollary}
\label{cor:euclid-hinge}
Let $(X_i,Y_i)$ be an iid sample of size $n$, where each $X_i\in\R^N$ satisfies $\nrm{X_i}_2\le1$.
Then for all $\delta>0$,
with probability at least $1-\delta$, 
for every $w\in\R^N$ with $\nrm{w}_2\le1$, 
and every $k$-dimensional subspace $T$ 
to which the sample is $\distor$-close,
we have
\beq
\E[L(w\cdot X,Y)]
\le 
\oo n\sum_{i=1}^n L(w\cdot X_i,Y_i)
+34\sqrt\frac{k}{n} + 2\sqrt\frac{\distor}{n}
+3\sqrt{\frac{\log(2/\delta)}{2n}},
\eeq
where
$
L(u,y) = (1-uy)_+
$
is the hinge loss.
\end{corollary}
\bepf
Follows from
the Rademacher generalization bound 
\cite[Theorem 3.1]{mohri-book2012},
the complexity estimate in
Theorem \ref{thm:euclidean}, 
and an application of Talagrand's contraction lemma \citet{LedouxTal91}
 to incorporate the hinge loss.
\enpf
\begin{remark}
The expected loss is bounded in Corollary~\ref{cor:euclid-hinge} in terms 
of the empirical loss on the \emph{original sample points $\set{X_i}$};
we can easily bound it also in terms of the empirical loss on the 
\emph{projected points $\set{P_T(X_i)}$}, 
because the hinge-loss is $1$-Lipschitz, and thus 
\beqn
\label{eq:euclid-hinge-proj-sample}
  \abs{ \oo n \sum_{i=1}^n L(w\cdot X_i,Y_i) 
    -  \oo n \sum_{i=1}^n L(w\cdot X\prl_i,Y_i) }
  \leq \oo n \sum_{i=1}^n \nrm{X_i\prp}_2
  \leq \left( \oo n \sum_{i=1}^n \nrm{X_i\prp}_2^2 \right)^{1/2}
  \leq \sqrt{\distor}.
\eeqn
The linear classifier (choice of $w$) can therefore be computed
by considering either the original sample or the projected points.
\end{remark}

\hide{
Implicit in Corollary \ref{cor:euclid-hinge} is 
a tradeoff between dimensionality reduction and distortion. Algorithmically,
this tradeoff may be optimized using 
PCA. It suffices to compute the singular value decomposition once, with
runtime complexity $O(n^3+Nn^2)$ \citep{MR1417720}. 
Then for each $1\le k\le N$, we obtain 
the lowest-distortion
$k$-dimensional subspace $T^{(k)}$,
corresponding to the top $k$ singular values. 
We then choose the value $1\le k\le N$ which minimizes the generalization bound of 
Corollary \ref{cor:euclid-hinge}
and 
construct a low-dimensional linear classifier on the projected data $(P_T(x_1),y_1),\ldots,(P_T(x_n),y_n)$, 
which is ``lifted'' to $\R^N$.
Our generalization bound holds in the original space $\R^N$ even without the projection and lifting,
although heuristically we expect improved performance for norm-regularized classifiers since now $w$ is allowed to be large without ``wasting'' magnitude
on irrelevant dimensions.

As PCA is already employed heuristically as a denoising filtering step in the supervised 
classification setting \citep{DBLP:journals/jmlr/BiBEBS03, HA07, 
DBLP:journals/tsp/VarshneyW11}, Corollary \ref{cor:euclid-hinge} provides apparently the 
first rigorous theory for choosing the best cutoff for the PCA singular values. }

\section{Adaptive dimensionality reduction: Metric case}
\label{sec:general}

In this section we extend the statistical analysis of Section \ref{sec:warmup}
from Euclidean spaces to the general metric case.
Suppose 
$(\X,\rho)$ is a metric space and 
we receive the training sample 
$(X_i,Y_i)$, $i=1,\ldots,n$,
with $X_i\in\X$ and $Y_i\in\set{-1,1}$.
Following von Luxburg and Bousquet
\citet{DBLP:journals/jmlr/LuxburgB04} 
and Gottlieb et al.\
\citet{DBLP:journals/tit/GottliebKK14},
the classifier we construct will be a Lipschitz function
(whose predictions are computed via Lipschitz extension that in turn 
uses approximate nearest neighbor search)
--- but with the added twist of a dimensionality reduction preprocessing step.
To motivate this approach, let us recall some results from 
\citet{DBLP:journals/jmlr/LuxburgB04,DBLP:journals/tit/GottliebKK14}.
von Luxburg and Bousquet \citet{DBLP:journals/jmlr/LuxburgB04} made the simple but
powerful observation that $1$-Nearest-Neighbor ($1$-NN) classification is essentially equivalent
to computing a real-valued Lipschitz extension $h$ from the $\pm1$-labeled sample points,
and classifying test points by thresholding $h$ at $0$. 
This made the 1-NN classifier amenable 
to analysis by Rademacher complexity and fat-shattering dimension techniques.
In particular,
the $1$-NN classifier (and hence the Lipschitz extension it induces)
can be computed efficiently within any fixed additive precision,
formalized as follows.
\begin{theorem}[\cite{DBLP:journals/tit/GottliebKK14}]
\label{thm:GKK14}
Let $(\X,\rho)$ be a metric space, and fix $0 < \eps < \frac{1}{32}$.
For a sample $S$ consisting of $n$ labeled points 
$(x_1,y_1),\ldots,(x_n,y_n)\in\X\times\set{-1,1}$,
with $d=\ddim(\set{x_1,\ldots,x_n})$,
let $f^*:\X\to\R$ be a {\em Lipschitz extension} of $S$,
meaning that (a) $f^*(x_i)=y_i$ for all $i\in[n]$,
(b) $f^*$ is $L$-Lipschitz (c) there is no $f':\X\to\R$ satisfying (a) with
a Lipschitz constant smaller than $L$.
Then there is a function $\tilde f:\X\to\R$ such that
\bit
\item[(i)]
$\tilde f(x)$ can be evaluated at each $x\in\X$ in
time $2^{O(d)}\log n + \eps^{-O(d)}$, after an initial computation of
$(2^{O(d)} \log n + \eps^{-O(d)})n$ time;
\item[(ii)]
$|\tilde f(x)-f^*(x)| \le 2\eps$ for all $x\in\X$.
\eit
\end{theorem}
As shown ibid., 
the $\eps$-approximation causes a very mild degradation 
of the generalization bounds.
The exponential dependence 
of the runtime --- both the precomputation and the evaluation on new points ---
on $\ddim(S)$ stands to benefit a great deal from even a slight reduction in dimensionality.

The remainder of this section is organized as follows.
In Section \ref{sec:rad-bds},
we
formalize the notion of ``nearly'' low-dimensional data in a metric space
and discuss its implication for Rademacher complexity.
We say that
$S=\set{x_i}\subset\X$
is \emph{$(\distor,\D)$-elastic}
if there is a $\tilde S=\smset{\tilde x_i}\subset\X$
such that
$\oo n\sum_{i=1}^n\rho(x_i,\tilde x_i)\le\distor$
and $\ddim(\tilde S)\le \D$.
We can prove that if our data set is $(\distor,\D)$-{elastic},
then the Rademacher complexity it induces on Lipschitz functions
can be bounded in terms of $\distor$ and $\D$ alone
(Theorem \ref{thm:low-dim-savings}),
independently of the ambient dimension $\ddim(\X)$.
Similarly to the Euclidean case (Theorem~\ref{thm:euclidean}),
we then use in Section~\ref{sec:met-gen} these Rademacher complexity estimates 
to obtain data-dependent error bounds (Theorem~\ref{thm:rade-lip-margin}).

In Section \ref{sec:class-proc},
we describe how to convert our distortion-based statistical bounds 
(for the Rademacher complexity and generalization error)
into an effective classification procedure.
To this end, we develop
a novel bicriteria approximation algorithm
presented in Section \ref{sec:alg}.
Informally,
given a set $S\subset\X$ 
and a target doubling dimension $\D$,
our method efficiently computes a set $\tilde S$
with $\ddim(\tilde S)\approx \D$
and
approximately
minimal the distortion $\distor$.
In the sample-preprocessing step that is applied before classifying new points,
we iterate the bicriteria algorithm
to find a near-optimal tradeoff
between dimensionality and distortion.
Having found an $\tilde S$ achieving near-optimal 
tradeoff,
we apply on it the machinery of Theorem~\ref{thm:GKK14},
and exploit its low dimensionality 
for fast approximate nearest-neighbor search.

\subsection{Rademacher bounds}
\label{sec:rad-bds}

We begin by
obtaining complexity estimates
for Lipschitz functions in (nearly) doubling spaces. 
This was done in
\citet{DBLP:journals/tit/GottliebKK14}
in terms of the fat-shattering dimension, 
but here we obtain 
data-dependent
bounds
by direct control over the covering numbers.
The following ``covering numbers by covering numbers'' lemma is 
a variant of the classic 
estimate \citet{MR0124720}.
\begin{lemma}
\label{lem:cov-cov}
Let $\F_L$ be the collection of $L$-Lipschitz 
functions 
mapping
the metric space $(\X,\rho)$
to $[-1,1]$,
and endow $\F_L$ with the $L_\infty$ metric:
\beq
\nrm{f-g}_\infty = \sup_{x\in\X}\abs{f(x)-g(x)},
\qquad f,g\in\F_L.
\eeq
Then the covering numbers of $\F_L$ may be estimated in terms of 
the covering numbers of $\X$:
\beq
\calN(\eps,\F_L,\nrm{\cdot}_\infty) \le \paren{\frac{8}{\eps}}^{ \calN(\eps/2L,\X,\rho)} ,
\qquad \qquad
\forall \eps\in(0,1)
.
\eeq
Hence, for a space $(\X,\rho)$ with diameter 1,
\beq
\log \calN(\eps,\F_L,\nrm{\cdot}_\infty) 
\le 
\paren{\frac{4L
}{\eps}}
^{\ddim(\X)}
\log
\paren{\frac{8}{\eps}} ,
\qquad
\forall \eps\in(0,1)
.
\eeq
\end{lemma}

Equipped
with 
the
covering numbers estimate, 
we proceed to bound the Rademacher complexity of Lipschitz functions
on doubling spaces.%
\footnote{Analogous bounds were obtained by \citet{DBLP:journals/jmlr/LuxburgB04} in less explicit form.}

\begin{theorem}
\label{thm:rade-doubling}
Let $\F_L$ be the collection of $L$-Lipschitz $[-1,1]$-valued functions defined on a metric space 
$(S,\rho)$
with diameter $1$ and doubling dimension $\D$.
Then
$
\hat R_n(\F_L;S)
=
O\paren{\frac{L}{n^{1/(\D+1)}}}.
$
\end{theorem}

\bepf
Recall that $\nrm{f}_2\le\nrm{f}_\infty$ implies
$\calN(\eps,\F,\nrm{\cdot}_2)
\le
\calN(\eps,\F,\nrm{\cdot}_\infty)
$.
\hide{
Integrate[ 1/x^{(\D+1)/2},{x,a,2}]
}
Substituting the estimate in
Lemma \ref{lem:cov-cov} into Dudley's integral (\ref{eq:dudley-int}), we have
\beq
\hat R_n(\F_L;S) &\le&
\inf_{\alpha\ge0}\paren{4\alpha+12\int_\alpha^\infty\sqrt{
\frac{
\log\calN(t,\F_L,\nrm{\cdot}_\infty)
}{n}}dt}\\
&\le&
\inf_{\alpha\ge0}\paren{4\alpha+12\int_\alpha^2\sqrt{
\frac{
\paren{\frac{4L}{t}}^{\D}\log\paren{\frac{8}{t}}
}{n}}dt}\\
&\le&
\inf_{\alpha\ge0}\paren{4\alpha+12\int_\alpha^2\sqrt{
\frac{
\paren{\frac{4L}{t}}^{\D}\paren{\frac{8}{t}}
}{n}}dt}\\
&\le&
\inf_{\alpha\ge0}\paren{4\alpha+\frac{34(4L)^{\D/2}}{\sqrt n}\int_\alpha^2
\paren{\frac{1}{t}}^{(\D+1)/2}
dt}\\
&=&
\inf_{\alpha\ge0}\paren{4\alpha+
\frac{34(4L)^{\D/2}}{\sqrt n}
\paren{\frac{\D-1}{2}}\paren{\oo{\alpha^{(\D-1)/2}}-2^{-(d+1)/2}}
}\\
&=& 4^{\frac{\D-2}{\D+1}}\paren{(\D-1)K}^{\frac{2}{\D+1}}
+
K\paren{
\paren{
8^{-\frac{2}{\D+1}}
((\D-1)K)^{\frac{2}{\D+1}}}^{\frac{\D-1}{2}}
-2^{-(d+1)/2}}
\\
&\le&
8K^{\frac{2}{\D+1}}
+
\D K\paren{
K^{\frac{\D-1}{\D+1}}
-2^{-(d+1)/2}}=O(K^{\frac{2}{\D+1}}),
\eeq
where
\beq
K=
\frac{34(4L)^{\D/2}}{\sqrt n}
\paren{\frac{\D-1}{2}}.
\eeq
Thus,
$
\hat R_n(\F_L;S) = O\paren{\frac{L}{n^{1/(\D+1)}}}
$,
as claimed.
\enpf

We can now quantify the savings earned by a low-distortion dimensionality reduction.

\begin{theorem}
\label{thm:low-dim-savings}
Let $(\X,\rho)$ be a metric space with diameter $1$,
and consider the two $n$-point sets $S,\tilde S\subset\X$,
where $S$ is 
%$(\distor n^{\D/(\D+1)},\D)$-elastic 
$(\distor,\D)$-elastic 
with witness $\tilde S$.
Let $\F_L$ be the collection of all $L$-Lipschitz, $[-1,1]$-valued functions
on $\X$.
Then
$
\hat R_n(\F_L; S) =
O\paren{
{L(1/n^{1/(d+1)}+\distor)}
%\frac{L(1+\distor)}{n^{1/(\D+1)}}
}
.
$
\end{theorem}
\bepf
For $X_i\in S$ and $\tilde X_i\in\tilde S$, 
define $\delta_i(f) = f(X_i)-f(\tilde X_i)$.
Then
\beqn
\hat R_n(\F_L; S) 
&= & \E\sup_{f\in\F_L}\oo n\sum_{i=1}^n \sigma_i f(X_i)
=
\E\sup_{f\in\F_L}\oo n\sum_{i=1}^n \sigma_i (f(\tilde X_i)+\delta_i(f))
\nonumber\\
&\le& \hat R_n(\F_L; \tilde S) 
+ \E\sup_{f\in\F_L}\oo n\sum_{i=1}^n \sigma_i \delta_i(f).
\label{eq:radecomp}
\eeqn
Now 
the Lipschitz property 
and
our definition of distortion
imply that
\beq
\abs{
\sum_{i=1}^n\sigma_i\delta_i(f)
}
\le 
\sum_{i=1}^n\abs{\delta_i(f)}
\le
L\sum_{i=1}^n \rho(X_i,\tilde X_i)
\le
%L\distor n^{\D/(\D+1)}
Ln\distor,
\eeq
and hence
$
\E\sup_{f\in\F_L}\oo n\sum_{i=1}^n \sigma_i \delta_i(f) 
\le
%\frac{L\distor}{n^{1/(\D+1)}}
L\distor
.
$
The other term in (\ref{eq:radecomp}) is bounded by invoking
Theorem \ref{thm:rade-doubling}.
\enpf
\begin{remark}
Comparing the Rademacher estimate $O\paren{\frac{1}{n^{1/(\D+1)}}
+\eta
}$ 
from Theorem~\ref{thm:low-dim-savings}
with the bound $O(\sqrt{\D/n}+\sqrt{\distor/n})$ 
from the Euclidean case (Theorem~\ref{thm:euclidean}),
we see an exponential gap in the dependence on the dimension.
The gap's origin is the bound (\ref{eq:l2-cov-3tk}) for linear classifiers
compared with Lemma~\ref{lem:cov-cov} for Lipschitz functions 
(the latter estimate is essentially tight \cite{MR0124720}).
\end{remark}

\subsection{Generalization bounds}
\label{sec:met-gen}

For $f:\X\to[-1,1]$,
define
the {\em margin} of $f$
on the labeled example $(x,y)$
by 
$yf(x)$.
The \emph{$\gam$-margin loss}, $0<\gam<1$,
that $f$ incurs on 
$(x,y)$ is
$L_\gam(f(x),y)=
\min(\max(0,1-yf(x)/\gam),1),$
which charges a value of $1$ for predicting the wrong sign, i.e., $yf(x)<0$,
charges nothing for predicting correctly with confidence $yf(x)\ge\gam$,
and interpolates linearly between these regimes.
Since $L_\gam(f(x),y)\le\pred{yf(x)<\gam}$,
the sample's $\gam$-margin loss is at most 
its $\gam$-margin misclassification error.

\begin{theorem}
\label{thm:rade-lip-margin}
Let $\F_L$ be the collection of $L$-Lipschitz functions mapping 
a metric space $\X$ of diameter 1 to $[-1,1]$.
If the iid sample $(X_i,Y_i)\in\X\times\set{-1,1}$, 
$i=1,\ldots,n$,
is
%$(\distor n^{\D/(\D+1)},\D)$-elastic,
$(\distor ,\D)$-elastic,
then
for any $\delta>0$,
with probability at least $1-\delta$, the following holds for
all $f\in\F_L$
and all $\gamma\in(0,1)$:
\beq
\P(\sgn(f(X))\neq Y)
\le
\oo n\sum_{i=1}^n L_\gam(f(X_i),Y_i)
+O\paren{
\frac{L (1/n^{1/(d+1)}+\eta) }{\gamma}
 %\frac{L(1+\distor)}{\gamma n^{1/(\D+1)}}
+\sqrt{\frac{\log\log(L/\gam)}{n}}
+\sqrt{\frac{\log(1/\delta)}{n}}}
.
\eeq
\end{theorem}
\bepf
We invoke 
\cite[Theorem 4.5]{mohri-book2012}
to bound the
classification error in terms of sample margin loss and Rademacher complexity
and the latter is bounded via
Theorem \ref{thm:low-dim-savings}.
\enpf
\begin{remark}
The expected loss is bounded in Theorem~\ref{thm:rade-lip-margin} in terms 
of the empirical loss on the \emph{original sample points $\set{X_i}$};
we can easily bound it also in terms of the empirical loss on the 
\emph{perturbed points $\set{\tilde X_i}$}, 
%because the $\gam$-margin loss is $(1/\gam)$-Lipschitz, and thus 
because the loss  $L_\gam(f(\cdot),Y_i)$  is $(L/\gam)$-Lipschitz, and thus
\beqn
\label{eq:rade-lip-margin-perturbed-sample}
  \abs{ \oo n \sum_{i=1}^n L_\gam(f(X_i),Y_i)
    - \oo n \sum_{i=1}^n L_\gam(f(\tilde X_i),Y_i) } 
  \leq \oo n \sum_{i=1}^n \frac{L}{\gam}\rho(x_i,\tilde x_i)
  \le \frac{L}{\gam} \distor.
\eeqn
\end{remark}

\subsection{Classification procedure}
\label{sec:class-proc}
Theorem \ref{thm:rade-lip-margin} provides a statistical optimality criterion
for the dimensionality-distortion tradeoff. 
Unlike the Euclidean case, where the well-known
PCA optimized this tradeoff,
the metric case requires the
bicriteria approximation
algorithm which we describe in Section \ref{sec:alg}.
To recap,
given a set $S\subset\X$ 
and a target doubling dimension $\D$,
this algorithm efficiently computes a set $\tilde S$
with $\ddim(\tilde S)\approx \D$, which approximately
minimizes the distortion $\distor$.
We may
iterate this algorithm over all
$\D\in\set{1,\ldots,\log_2|S|}$ 
--- since the doubling dimension of the metric space $(S,\rho)$ is at most $\log_2|S|$
--- to optimize the complexity term%
\footnote{
Since $L/\gam$ multiplies the main term $1/n^{1/(\D+1)}+\distor$ in
the error bound, and the other term is of order $\log\log(L/\gam)$,
the optimization may effectively be carried out oblivious to $L$ and $\gam$.
} 
in Theorem \ref{thm:rade-lip-margin}.
 
Once a nearly optimal witness $\tilde S$ of
$(\distor,\D)$-elasticity has been constructed,
we predict the value at a test point $x\in\X$
by a thresholded Lipschitz extension from $\tilde S$,
as provided in Theorem~\ref{thm:GKK14}.

\section{Approximately Optimizing the Dimension-Distortion Tradeoff}
\label{sec:alg}
In this section we cast the computation
of an $(\distor,\D)$-elasticity witness of a given finite set
as an optimization problem
and design for it a polynomial-time bicriteria approximation algorithm.
Let $(\X,\rho)$ be a finite metric space.
For a point $v$ and a point set $T$, 
define $\rho(v,T) = \min_{w \in T} \rho(v,w)$. 
Given two point sets $S,T$, 
define the {\em cost of mapping} $S$ to $T$ to be $\sum_{v \in S} \rho(v,T)$. 
Define the \emph{Low-Dimensional Mapping (LDM)} problem as follows: 
Given a point set $S\subseteq \X$ and a target dimension $\D\ge1$, 
find $T\subseteq S$ with $\ddim(T) \le \D$ 
such that the cost of mapping $S$ to $T$ is minimized.%
\footnote{The LDM problem differs from $k$-median (or $k$-medoid) in 
that it imposes a bound on $\ddim(T)$ rather than on $|T|$.
}
An $(\alpha,\beta)$-bicriteria approximate solution to the LDM problem 
is a subset $V \subset S$,
such that the cost of mapping $S$ to $V$ is at most $\alpha$ times 
the cost of mapping $S$ to an optimal $T$ (of $\ddim(T)\leq \D$), 
and also $\ddim(V) \le \beta \D$. 
We prove the following theorem.

\begin{theorem}\label{thm:bicriteria}
The Low-Dimensional Mapping problem admits 
an $(O(1),O(1))$-bicriteria approximation in runtime 
$2^{O(\ddim(S))}n + O(n \log^4n)$, where $n=|S|$.
\end{theorem}

The presentation of the algorithm proceeds in four steps.
In the first step, we modify LDM by adding to it yet another constraint:
We simply extract from $S$ a point hierarchy $\SS$ (see Section
\ref{sec:hier}), and require that the LDM solution not only be a subset of 
$S$, but that it also possesses a hierarchy which is a
sub-hierarchy of $\SS$. Lemma \ref{lem:hier} demonstrates
that this additional requirement can be fulfilled without
significantly altering the cost and dimension of the optimal
solution.

The second step of the presentation is an integer program (IP) which models 
the modified LDM problem, in Section \ref{sec:ip}.
We show that a low-cost solution to the LDM problem implies 
a low-cost solution to the IP, and vice-versa (Lemma \ref{lem:ip}). 

Unfortunately, finding an optimal solution to the IP seems difficult, hence
the third step, in Section \ref{sec:lp}, is to relax some of the IP constraints
and derive a linear program (LP). We also give a rounding scheme that
recovers from the LP solution an integral solution, 
and show that this integral solution indeed provides 
an $(O(1),O(1))$-bicriteria approximation (Lemma \ref{lem:lp}).
The final step, in Section \ref{sec:solver}, shows that the LP can be 
solved in the runtime stated above (Lemma \ref{lem:solver}),
thereby completing the proof of Theorem \ref{thm:bicriteria}.

\begin{remark}
The presented algorithm has very large (though constant) approximation factors.
The introduced techniques can yield much tighter bounds, 
by creating many different point hierarchies instead of only a single one. 
We have chosen the current presentation for simplicity.
\end{remark}

\subsection{Point hierarchies} \label{sec:hier}

Let $S$ be a point set, and assume by scaling that it has diameter $1$ 
and minimum interpoint distance $\delta>0$. 
A {\em hierarchy} $\SS$ of a set $S$ is a
sequence of nested sets $S_0 \subseteq \ldots \subseteq S_t$;
here, $t=\ceil{\log_2 (1/\delta)}$ and $S_t = S$, while $S_0$ consists of a single
point. 
Set $S_i$ must possess a \emph{packing} property, 
which asserts that $\rho(v,w) \ge 2^{-i}$ for all $v,w \in S_i$, 
and a \emph{$c$-covering} property for $c\ge 1$ (with respect to $S_{i+1}$),
which asserts that for each $v \in S_{i+1}$ there exists $w \in S_i$ with
$\rho(v,w) < c\cdot 2^{-i}$. 
Set $S_i$ is called a \emph{$2^{-i}$-net} of the hierarchy.
We remark that every point set $S$ possesses, for every $c\geq 1$, 
at least one hierarchy.

We modify LDM by adding to it yet another constraint.
Given set $S$ with 1-covering hierarchy $\SS$, the solution must also possess a
hierarchy which is a sub-hierarchy of $\SS$. 
(This implies a more structured solution, for which we can construct
an IP in Section \ref{sec:ip}.) We show
that this additional requirement can be fulfilled without
significantly altering the cost and dimension of the optimal
solution.

\begin{lemma}\label{lem:hier}
Let $S$ be a point set, and let $\SS=(S_0,\ldots,S_t)$ 
be a hierarchy for $S$ with a $1$-covering property. For every subset
$T \subset S$ with doubling dimension $\D := \ddim(T)$, there exists a set $V$ 
satisfying $T \subseteq V \subseteq S$, and
an associated hierarchy $\V=(V_0,\ldots,V_t)$ with the following properties.
\begin{enumerate}
\compactify
\item Dimension: $\ddim(V) \leq \Dhat :=  4\D + 1$.
\item Covering: Every point $v \in V_i$ is 
$4$-covered by some point in $V_{i-1}$, and 
$5$-covered by some point of $V_{k}$ for all $k<i$.
\item Heredity: $\V$ is a \emph{sub-hierarchy} of $\SS$, 
meaning that $V_i \subseteq S_i$ for all $i\in[t]$. %
\end{enumerate}
\end{lemma}

\bepf
First extract from the set $T$ an arbitrary $1$-covering hierarchy 
$\TT=(T_0,\ldots,T_t)$. Note that 
each point $v \in T_i$ is necessarily within distance $2 \cdot 2^{-i}$ of some point in $S_i$; 
this is because $v\in S_t$, and by the $1$-covering property of $\SS$, 
there must be some point $w \in S_i$ within distance 
$\sum_{j=i}^{t} 2^{-j} 
< 2 \cdot 2^{-i}$. 

Initialize the hierarchy $\V$ by setting $V_0 = S_0$. 
Construct $V_i$ for $i>0$ by first including in 
$V_i$ all points of $V_{i-1}$. 
Then, for each $v \in T_i$ (in an arbitrary order), 
if $v$ is not within distance $2 \cdot 2^{-i}$ of a 
point already included in $V_i$, add to $V_i$ the point $v' \in S_i$ closest to $v$.
(Recall from above that $\rho(v,v') < 2 \cdot 2^{-i}$.)
Let the final set $V_t$ also include all points of $T_t$
-- that is all of $T$ -- and $V = V_t$.

Observe that $\V$ is a sub-hierarchy of 
$\SS$, thus it inherits the packing property of hierarchy $\SS$.
Further, since $\TT$ obeyed a $1$-cover property, every point in $T_i$ is within distance 
$2^{-i+1}$ of some point in $T_{i-1}$, 
and so the scheme above implies that any point in $V_i$ 
must be within distance
$2 \cdot 2^{-i} +  2^{-i+1} + 2 \cdot 2^{-(i-1)} 
= 4 \cdot 2^{-i+1}$
of some point in $V_{i-1}$.
(Here, we have used the triangle inequality from $V_i$ 
to $T_i$, $T_i$ to $T_{i-1}$ and finally $T_{i-1}$ to $V_{i-1}$.)
Likewise, since every point in $T_i$ is, for any $k<i$, 
within distance $2 \cdot 2^{-k}$ of some point in $T_k$, 
we get that any point in $V_i$ must be within distance
$2 \cdot 2^{-k} + 2 \cdot 2^{-i} + 2 \cdot 2^{-k}
\le 5 \cdot 2^{-k}$
of some point in $V_k$.

Turning to the dimension, consider an arbitrary ball of radius $2^{-i}$ 
centered at any point $v \in V$. Let $B$ be the set of points of $V$ contained
in the ball; we show that $B$ can be covered by a small number of balls
of radius $2^{-i-1}$:

Define $B' \subseteq B$ to include point of $V$ found only in levels $V_k$ for $k \ge 
i+3$. $B'$ can be covered by balls of radius $2^{-i-1}$ 
centered at points of $T_{i+3}$: By construction, any point first appearing in $V_{i+3}$ 
is within distance $2 \cdot 2^{-i-3} <  2^{-i-1}$ of a point of $T_{i+3}$. Any 
point first appearing in level $V_k$ for $k>i+3$ is within distance $2 \cdot 2^{-k}$ of a 
point in $T_k$, and that point is within distance $2 \cdot 2^{-i-3}$ of its ancestor in 
$T_{i+3}$, for a total distance of $2 \cdot 2^{-k} + 2 \cdot 2^{-i-3} < 2^{-i-1}$. 
It follows that all points of $B'$ can be covered by a set of balls of radius 
$2^{-i-1}$ centered at points of $T_{i+3}$ within distance 
$2^{-i} + 2^{-i-1} < 2^{-i+1}$ of $v$. By the doubling property of $T$, these account for at 
most $2^{4\D}$ balls.

Now consider the remaining points $B\setminus B'$. These points are found in $V_k$ for
$k \le i+2$. When $k \le i$, by construction $V_k$ possesses minimum interpoint 
distance $2c \cdot 2^{-i}$, and so only a single point among all these levels 
may be found in $B$. When $k = i+1$, we charge each point of 
$V_k$ to its nearest point in $T_k$, and note that the points of $T_k$ are within
distance $2 \cdot 2^{-k} + 2^{-i} = 4 \cdot 2^{-k}$ of $v$. There can be at most
$4^{\D}$ such points. A similar argument gives fewer than $8^\D$ points when $k = i+2$.
It follows that $B$ can be covered by $2^{4\D} + 1 + 4^\D + 8^\D < 2^{4\D+1}$ smaller 
balls.
\enpf

\subsection{An integer program}\label{sec:ip}

We now show an integer program%
\footnote{Formally, this is a mixed integer linear program,
since only some of the variables are constrained to be integers.
}
that encapsulates a near-optimal solution to the modified LDM problem;
we later relax it to a linear program in Section \ref{sec:lp}. 
Denote the input by $S=\{v_1,\ldots,v_n\}$ and the target dimension by $\D\ge1$,
and let $\SS$ be a hierarchy for $S$ with a $1$-covering property. 
We shall assume, following Section \ref{sec:hier}, 
that all interpoint distances in $S$ are in the range $[\delta,1]$,
and the hierarchy possesses $t = \lceil \log_2(1/\delta) \rceil$ levels.
The optimal IP solution will imply a subset $W \subset S$
equipped with a hierarchy $\W=(W_0,\ldots,W_t)$ that is a sub-hierarchy of $\SS$; 
we will show in Lemma~\ref{lem:ip} that $W=W_t$ constructed in this way 
is indeed a bicriteria approximation to the modified LDM problem,
and therefore to the original LDM problem as well.

We introduce a set $Z$ of 0-1 variables for the hierarchy $\SS$;
each variable $z^i_j \in Z$ corresponds to a point $v_j \in S_i$,
so clearly $|Z| \le nt$.
The IP imposes in Constraint \eqref{con:ip-integer} that $z^i_j \in \{0,1\}$,
and this variable is intended to be an indicator for whether $v_j\in W_i$.
The IP requires in Constraint \eqref{con:ip-nested} that $z^i_j \le z^{i+1}_j$,
to enforce that the hierarchy $\W$ is nested,
i.e., if a point is present in $W_i$ then it should be present also in $W_{i+1}$,
and in fact also in $W_k$ for all $k>i$.
When convenient, we may refer to distance between variables
where we mean distance between their corresponding points 
(i.e., distance from $z^i_j$ actually means distance from $v_j$).

We shall now define the $i$-level {\em neighborhood} of a point $v_j$ to be 
the points in $S_i$ that
are relatively close to $v_j$, using a parameter $\alpha\ge1$. 
Formally, when $v_j \in S_i$, 
let $N^i_j (\alpha) \subseteq Z$ include all variables
$z^i_k$ for which $\rho(v_j,v_k) \le \alpha \cdot 2^{-i}$,
and call $v_j$ the \emph{center} of $N^i_j (\alpha)$.
If $v_j \notin S_i$, then let $w \in S_i$ be the nearest
neighbor of $v_j$ in $S_i$ (recall that $\rho(v_j,w) < 2 \cdot 2^{-i}$), 
call $w$ the \emph{center} of $N^i_j (\alpha)$,
and define $N^i_j (\alpha) \subseteq Z$ to include all variables
$z^i_k$ for which $\rho(w,v_k) \le \alpha \cdot 2^{-i}$. 
So in both cases, $N^i_j (\alpha)$ is the set of variables $z^i_k$ 
within distance $\alpha\cdot2^{-i}$ from the center.
Constraint \eqref{con:ip-pack} of the IP 
bounds the number of points that $W$ includes 
from the set $N^i_j (\alpha)$ for each $\alpha\in\aset{7,24,75,588,612}$;
by the packing property 
for doubling spaces of dimension $\Dhat:= 4\D +1$ (Lemma \ref{lem:hier})  
it can be formulated as 
$\sum_{z \in N^i_j(\alpha)} z 
\le \lceil 2^{\log \alpha} \rceil^{\Dhat}
\le (2 \alpha)^\Dhat $.
The IP also imposes a covering property, 
by requiring in Constraint \eqref{con:ip-cover} that 
$\sum_{z \in N^i_j (7)} z \ge z^{t}_j$. 
Because $S$ possesses a $1$-covering property,
this constraint ensures indirectly that any point in $W_{i+1}$ is
$8$-covered by some point in $W_i$
(recall $N^i_j(\alpha)$ might be defined via the nearest neighbor in $S_i$),
and $9$-covered by some point in $W_k$ for all $k<i$.

We further introduce in Constraint~\eqref{con:ip-cost0} a set $C$ of $n$ cost variables $c_j\ge0$
to represent the point mapping cost $\rho(v_j,W)$,
and these are enforced by Constraints \eqref{con:ip-cost1}-\eqref{con:ip-cost2}.

The complete integer program is as follows.
{ %
\begin{align}
    \text{min}
    & \quad \sum_j c_j  \nonumber \\
    \text{s.t.}
    & \quad z^i_j \in \{0,1\} 
    & \forall z^i_j\in Z 
        \label{con:ip-integer}  \\
    & \quad z^i_j \le z^{i+1}_j
    & \forall z^i_j\in Z,\ i<t
        \label{con:ip-nested}  \\
    & \quad \sum_{z \in N^i_j(\alpha)} z \le (2\alpha)^{\Dhat}
    & \forall \alpha\in\aset{7,24,75,588,612},\ i\in[0..t],\ v_j\in S
        \label{con:ip-pack} \\
    & \quad \sum_{z \in N^i_j(7)} z \ge z^{t}_j
    & \forall i\in[0..t],\ \forall z^t_j\in Z
        \label{con:ip-cover}        \\
    & \quad \sum_{z \in N^i_j(24)} z \ge \tfrac{1}{(2 \cdot 24)^{\Dhat}}\sum_{z \in N^k_j(24)} z
    & \forall i,k\in[0..t], i<k,\ \forall v_j\in S 
        \label{con:ip6} \\
    & \quad c_j\ge 0
    & \forall v_j\in S 
        \label{con:ip-cost0} \\
    & \quad  z^t_j + \tfrac{c_j}{\delta} \ge 1
    & \forall v_j\in S
        \label{con:ip-cost1} \\
    & \quad  z^t_j + \tfrac{c_j}{2^{-i}} + \sum_{z \in N^i_j(12)} z \ge 1
    & \forall i\in[0..t],\ \forall v_j\in S
        \label{con:ip-cost2} 
\end{align}
}

Recall that $T$ denotes an optimal solution for the original low-dimensional mapping problem
on input $(S,\D)$, and let $C^*$ be the cost of mapping $S$ to $T$. Let $V$ be the set
guaranteed by Lemma \ref{lem:hier} (applied with covering $c=1$), 
and since $T \subseteq V$ the cost of mapping $S$ to $V$ is no
greater than $C^*$. The following lemma proves a bi-directional relationship 
between the IP and LDM, 
to eventually bound the IP solution $W$ in terms of the LDM solution $T$.
We remark that Constraint \eqref{con:ip6} is not necessary for the following
lemma, but will later play a central role in the proof of Lemma \ref{lem:lp}.

\begin{lemma}\label{lem:ip}
Let $(S,\D)$ be an input for the LDM problem,
and let $T\subseteq V$ be solutions as above. Then
\begin{enumerate} \compactify
\item[(a)]
$V$ implies a feasible solution to the IP of objective value at most $C^*$.
\item[(b)]
A feasible solution to the IP with objective value $C'$ yields (efficiently) 
an LDM solution $W$ with 
$\ddim(W) \leq \log 150 \cdot \Dhat < 29 \D + 8$ 
and cost of mapping $S$ to $W$ at most $32 C'$. 
\end{enumerate}
\end{lemma}

\bepf
To prove part (a), we need to show that assigning the indicator variables in $Z$ and 
cost variables $C$ according to $V$ yields a feasible solution with the stated 
objective value.
Note that $\V$ is nested, so it satisfies Constraint \eqref{con:ip-nested}.
Further, the doubling dimension of $V$ (as given in Lemma \ref{lem:hier})
implies that all points obey packing Constraint \eqref{con:ip-pack}.
The covering properties of $V$ are in fact tighter than those required by Constraint 
\eqref{con:ip-cover}.

Turning to the IP objective value, 
let us show that setting $c_j=\rho(v_j,V)$
satisfies the cost Constraints \eqref{con:ip-cost1}-\eqref{con:ip-cost2}.
Verifying Constraint \eqref{con:ip-cost1} is easy;
we assume that $c_j<\delta$ (as otherwise the constraint is clearly satisfied),
and then necessarily $v_j\in V$ and thus $z^t_j=1$.
To verify Constraint \eqref{con:ip-cost2} for a given $i$,
we may assume that $c_j<2^{-i}$ (otherwise the constraint is clearly satisfied). 
Let $v^*$ be the closest neighbor to $v_j$ in $V$.
The point $v^*$ is $9$-covered by some point $v_p\in V_i$ 
(using Constraint \eqref{con:ip-cover}, just as we explained about $\W$), 
and so
$\rho(v_j,v_p)
\le \rho(v_j,v^*) + \rho(v^*,v_p)
\le 2^{-i} + 9 \cdot 2^{-i}
= 10 \cdot 2^{-i}$.
Now, the distance from $v_j$ to the closest point in $S_i$ is less than $2 \cdot 2^{-i}$, 
so $v_p$ is within distance
$10 \cdot 2^{-i} + 2 \cdot 2^{-i} = 12 \cdot 2^{-i}$ of the center of $N^i_j(12)$.
We thus find a variable $z^i_p$ included in $N^i_j(12)$ that has value $1$ 
(recall $v_p\in V_i$).

We next claim that Constraint \eqref{con:ip6} is actually extraneous for this IP, and follows
from Constraint \eqref{con:ip-cover}:
Constraint \eqref{con:ip6} simply means that if $N^k_j(24)$ contains at least one non-zero variable, then 
so does $N^i_j(24)$. (Recall that by Constraint \eqref{con:ip-pack}, 
$N^k_j(24)$ contains at most $(2 \cdot 24)^{\Dhat}$ non-zero variables.)
But if $N^k_j(24)$ contains a non-zero variable, then this variable is necessarily 
$9$-covered by some non-zero variable in level $i$ of hierarchy $\V$ (by Constraint \eqref{con:ip-cover}). 
The covering variable must be within distance 
$9 \cdot 2^{-i} +  24 \cdot 2^{-k}$ of the center of $N^k_j(24)$, within distance
$9 \cdot 2^{-i} +  24 \cdot 2^{-k} + 2 \cdot 2^{-k} \le 22 \cdot 2^{-i} $ of $v_j$,
and within distance $22 \cdot 2^{-i} + 2 \cdot 2^{-i} = 24 \cdot 2^{-i}$ of the center of 
$N^i_j(24)$.
So $N^i_j(24)$ contains the covering non-zero variable.

To prove part (b), given a solution to the IP, let the indicator variables $Z$
determine the points of LDM solution $W \subset S$, as well as its hierarchy.
We need to bound the dimension and the mapping cost of $W$. 
Consider a ball of radius $2^{-i} < r \le 2^{-i+1}$ centered at $v_j \in W$, 
and let us show that it can be covered by a bounded number of balls of radius
at most $2^{-i-1}<\frac{r}{2}$.
Every point covered by $v_j$'s ball is within distance 
$9 \cdot 2^{-i-5} < 2^{-i-1}$ of some covering point in $W_{i+5}$:
To bound the number of these covering points in $W_{i+5}$,
observe they are all within distance 
$9 \cdot 2^{-i-5} + r
\le 9 \cdot 2^{-i-5} + 2^{-i+1}  
= 73 \cdot 2^{-i-5}$ 
of $v_j$, 
and within distance $75 \cdot 2^{-i-5}$ of a point $v \in S_{i+5}$ covering $v_j$,
and that by Constraint \eqref{con:ip-pack}, 
there are at most $(2 \cdot 75)^{\Dhat}$ net-points of $W_{i+5}$ within distance 
$75 \cdot 2^{-i-5}$ of $v$.
It follows that $v_j$'s ball can be covered by $(2 \cdot 75)^{\Dhat}$ balls
of radius $9 \cdot 2^{-i-5} < 2^{-i-1}$ (and whose centers are in $W$), 
and therefore $\ddim(W) \leq \log 150 \cdot \Dhat$.

Turning to the mapping cost, we will demonstrate that $\rho(v_j,W) \le 32c_j$. 
If $\rho(v_j,W)=0$ the claim holds trivially,
so we may assume $2^{-p} \le \rho(v_j,W) < 2^{-(p-1)}$ for some integer $p$.
If $p\ge t-4$ then using Constraint \eqref{con:ip-cost1} we have that
$c_j \ge \delta \ge 2^{-t} \ge 2^{-(p-1)-5} > \frac{1}{32}\rho(v_j,W)$.
Otherwise, we shall use Constraint \eqref{con:ip-cost2} for $i=p+5\le t$.
The distance from $v_j$ to any point of $N^i_j(24)$ 
is at most $2 \cdot 2^{-i} + 24 \cdot 2^{-i} = 26 \cdot 2^{-i} < 2^{-p}$,
but since $\rho(v_j,W) \geq 2^{-p}$, 
no point of $N^i_j(24)$ is contained in $W$
and thus $z^t_j = \sum_{z \in N^i_j(24)} z = 0$.
Constraint \eqref{con:ip-cost2} now implies 
$c_j \geq 2^{-i} = 2^{-p-5} \ge \frac{1}{32} \rho(v_j,W)$.
\enpf

\subsection{A linear program}\label{sec:lp}

While the IP gives a good approximation to the LDM problem, 
we do not know how to solve this IP in polynomial time.
Instead, we create an LP by relaxing the integrality constraints
\eqref{con:ip-integer} into linear constraints $z^i_j \in [0,1]$. 
This LP can be solved quickly, as shown in Section \ref{sec:solver}.
After solving the LP, we recover a solution to the LDM problem 
by rounding the $Z$ variables to integers, as follows:
\begin{enumerate}
\compactify
\item \label{it:s1}
If $z^t_j \ge \frac{1}{2}$, then $z^t_j$ is rounded up to 1.
\item \label{it:s2}
For each level $i=0,\ldots,t$:
Let $\hN^i$ be the set of all neighborhoods $N^i_j(24)$.
Extract from $\hN^i$ a maximal subset ${\hN}^i$ whose elements obey the following:
(i) For each $N^i_j(24) \in {\hN}^i$ there is some $k \ge i$ such that
$\sum_{z \in N^k_j(24)} z \ge \frac{1}{4}$.
(ii) Elements of ${\hN}^i$ do not intersect.
For each element $N^i_j(24) \in {\hN}^i$, we round up to $1$ its center $z^i_l$ 
(where $v_l$ is the nearest neighbor of $v_j$ in $S_i$),
as well as every variable $z^k_l$ with $k>i$.
\item \label{it:s3}
All other variables of $Z$ are rounded down to $0$.
\end{enumerate}

These rounded variables $Z$ correspond (in an obvious manner) 
to an integral solution $W'$ with hierarchy $\W'$. 
The following lemma 
completes the first half of Theorem \ref{thm:bicriteria}.

\begin{lemma}\label{lem:lp}
$W'$ is a $(624,82 + o(1))$-bicriteria approximate solution to 
the LDM problem on $S$.
\end{lemma}

\bepf
Before analyzing $W'$, we enumerate three properties of its hierarchy $\W'$.

\smallskip
\noindent {\em (i) Nested.} When a variable of level $i$ is rounded up in rounding 
step \ref{it:s2}, all corresponding variables in
levels $k>i$ are also rounded up. This implies that $\W'$ is nested.

\smallskip
\noindent {\em (ii) Packing.}
For later use, we will need to show that after the rounding, 
the number of 1-valued variables found in each $N^i_j(g)$ is small, when $g= 588$.
By Constraint \eqref{con:ip-pack}, the sum of the pre-rounded variables $z^i_k \in N^i_j(g)$ is at most 
$(2g)^{\Dhat}$.
If $i=t$, then step \ref{it:s1} rounds up only variables $z^t_k$ of value $\frac{1}{2}$ and higher, 
so after this rounding step $G^t_j$ contains at most $2 \cdot (2g)^{\Dhat}$ points of $W_t'$.
For general $i\in[t]$, variables of $N^i_j(g)$ may be rounded up 
due to rounding step 2 acting on level $i$.
This step stipulates that a variable $z^i_l \in N^i_j(g)$ may be rounded up if $z^i_l$ is the center of
a distinct subset $N^i_l(24) \in {\hN}^i$. Inclusion in ${\hN}^i$ requires
$\sum_{z \in N^k_l(24)} z \ge \frac{1}{4}$ for some $k \ge i$, and 
so Constraint \eqref{con:ip6} implies that
$\sum_{z \in N^i_l(24)} z \ge \frac{1}{4 (2 \cdot 24)^{\Dhat}}$. 
Now, since $z^i_l$ is in both $N^i_j(g)$ and $N^i_l(24)$, all points in $N^i_l(24)$ are within distance 
$g+24$ of the center of $G^i_j$, and so by Constraint \eqref{con:ip-pack} 
rounding step 2 may place at most
$4 (2 \cdot 24)^{\Dhat} \cdot (2 \cdot (g+24))^{\Dhat} <
(2 \cdot (g+24))^{2\Dhat}$ points of 
$W_i'$ into the ball.

Further, rounding step \ref{it:s2} acting on levels $k<i$ may add points to ball $N^i_j(g)$. Since points 
in each nested level $k$ possess packing $2^{-k}$, and the radius of our ball is at most $g \cdot 2^i$, 
levels $k \le i- \log g$ can together add just a single point. Levels 
$i-\log g<k<i$ may each add at most $(2 \cdot (g+24))^{\Dhat}$ additional points to $N^i_j(g)$,
accounting for $(2 \cdot (g+24))^{\Dhat}$ total points. It follows that
the total number of points in the ball is bounded by $2(2 \cdot (g+24))^{2\Dhat}$.

\smallskip
\noindent {\em (iii) Covering.} 
We first consider a variable $z^t_j$ rounded up in rounding step \ref{it:s1}, 
and show it will be $74$-covered in each level $W_i'$ of the hierarchy. 
Since $z^t_j \ge \frac{1}{2}$, 
Constraint~\eqref{con:ip-cover} implies that for the pre-rounded variables, 
$\sum_{z \in N^i_j(24} z \ge \sum_{z \in N^i_j(7)} z \ge 
\frac{1}{2}$. By construction of rounding step \ref{it:s2}, a variable of 
$N^i_j(24)$ or one in a nearby set in $\hN^i$ is 
rounded to 1, and the distance of this variable from $v_j$ is less than 
$(3 \cdot 24 +2)\cdot 2^{-i} = 74 \cdot 2^{-i}$.

We turn to a variable $z^i_j$ rounded to 1 in step \ref{it:s2}, 
and demonstrate that it too is $74$-covered in each hierarchy level $k<i$. 
Since $z^i_j$ was chosen to be rounded, there must exists $k \ge i$ 
with $\sum_{z \in N^k_j(24)} z \ge \frac{1}{4}$, and so a variable in every set
$N^h_j(24)$ (or in a nearby set in $\hN^h$) for all $h<k$ must be rounded as well.
It follows that $z^i_j$ is $3 \cdot 24 < 74$-covered by a variable in each set $N^h_j(24)$
(or in a nearby set in $\hN^h$) for all $h<i$.

\medskip

Having enumerated the properties of the hierarchy, we can now prove the doubling dimension of $W'$.  
First fix a radius $2^{-i} < r \le 2^{-i+1}$ and a center $v_j \in W_t'$, and we
will show that this ball can be covered by a fixed number of balls with radius
at most $2^{-i-1} < \frac{r}{2}$: Each point covered by $v_j$'s ball is within distance 
$74 \cdot 2^{-i-8} < 2^{-i-1}$ of some covering point in $W_{i+8}$, and so all covering
points in $W_{i+8}$ are within distance 
$74 \cdot 2^{-i-8} + 2^{-i+1} 
= 74 \cdot 2^{-i-8} + 512 \cdot 2^{-i-8}
= 586 \cdot 2^{-i-8}$ 
of $v_j$ and within distance $588 \cdot 2^{-i-8}$ of a point $v \in S_{i+8}$ covering $v_j$.
Recalling that $g=588$, by the packing argument above
there are at most $2(2 \cdot (g+24))^{2\Dhat}$ net-points of $W_{i+8}$ within distance 
$g \cdot 2^{-i-8}$ of $v$, and this implies a
doubling dimension of 
$2\Dhat \log(2 \cdot (g+24)) + 1 = 2\Dhat \log 1224 + 1 < 82\D + 1$.

It remains to bound the mapping cost. 
By Lemma \ref{lem:ip}(a), the cost of an optimal LP solution is at most $32C^*$.
Consider the mapping cost of a point $v_j$. 
If the corresponding variable $z^t_j$ was rounded up to $1$
then the mapping cost $\rho(v_j,W') = 0 \le c_j$,
i.e., at most the contribution of this point to the LP objective.
Hence, we may restrict attention to a variable $z^t_j<\frac{1}{2}$ 
that was subsequently rounded down. 
We want to show that $\rho(v_j,W')$ is not much more than the LP cost $c_j$. 
First, $c_j \ge \frac{\delta}{2}$ by Constraint \eqref{con:ip-cost1}. 
Now take the highest level $i$ for which $c_j < \frac{2^{-i}}{4}$;  
by Constraint \eqref{con:ip-cost2}, it must be that
$ \sum_{z \in N^i_j(24)} \ge \frac{1}{4}$. 
Then by rounding step \ref{it:s2}, a variable within distance
$72+2 \cdot 2^{-i} = 74 \cdot 2^{-i}$ of $v_j$ must be rounded up. 
Hence, the LP cost $c_j \ge \frac{2^{-i-1}}{4} = \frac{2^{-i}}{8}$ 
is at least $1/592$-fraction of the mapping cost $\rho(v_j,W')$. 
Altogether, we achieve an approximation of $32 + 592 = 624$ to the optimal cost.
\enpf

\subsection{LP solver}\label{sec:solver}

To solve the linear program, we utilize the framework presented by Young \cite{Y01} for LPs of following form: Given 
non-negative matrices $P,C$, vectors $p,c$ and precision $\beta>0$, find a non-negative vector $x$ such that $Px 
\le p$ (LP packing constraint) and $Cx \ge c$ (LP covering constraint). Young shows that if there exists a 
feasible solution to the input instance, then a solution to a relaxation of the input program, specifically $Px 
\le (1+\beta)p$ and $Cx \ge c$, can be found in time $O(mr (\log m)/\beta^2)$, where $m$ is the number of 
constraints in the program and $r$ is the maximum number of constraints in which a single variable may appear. We 
will show how to model our LP in a way consistent with Young's framework, and obtain an algorithm that achieves 
the approximation bounds of Lemma \ref{lem:lp} with the runtime claimed by Theorem \ref{thm:bicriteria}. Lemma 
\ref{lem:solver} below completes the proof of Theorem \ref{thm:bicriteria}.

\begin{lemma}\label{lem:solver}
An algorithm realizing the bounds of Lemma \ref{lem:lp} can be computed in time $2^{O(\ddim(S))}n + O(n \log^4n)$.
\end{lemma}

\bepf
To define the LP, we must first create a hierarchy for $S$, which can be done in time
$\min \{ O(tn^2),2^{O(\ddim(S))}tn \}$, as in \citep{KL04,CG06}. 
After solving the LP, the 
rounding can be done in this time bound as well.

To solve the LP,
we first must modify the constraints to be of the form $Px \le p$ and $Cx \ge c$. This can be done easily by introducing
complementary constraints $\bar{z}^i_j \in [0,1]$, and setting $z^i_j + \bar{z}^i_j = 1$. 
For example, constraint
$z^i_j \ge z^t_j$ now becomes $z^i_j + \bar{z}^t_j \ge 1$. A similar approach works for the other constraints as well.

We now count the number of basic constraints. Note that $j \in [1,n]$ and $i \in [1,t]$, so a simple count gives
$m=O(t^2n)$ constraints (where the quadratic term comes from constraint \eqref{con:ip6}).
To bound $r$, the maximum number of constraints in which a single variable may appear, we note that this can 
always be bounded by $O(1)$ if we just make copies of variable $z^i_j$. (That is, two copies of the form
${z^i_j}' = z^i_j$, ${z^i_j}'' = z^i_j$, then two copies of each copy, etc.) So $r=O(1)$ and the bound on
$m$ increases to $O(t^2 n + n \log n)$.

Finally, we must choose a value for $\beta$. The variable copying procedure above creates a dependency chain of $O(\log n)$
variables, which will yield additive errors unless $\beta = O(1/\log n)$. Similarly, constraint \eqref{con:ip-nested} 
creates a chain of 
$O(t)$ variables, so $\beta = O(1/t)$. It suffices to take $\beta = O(1/(t\log n))$, and the stated runtime follows.
\enpf

\section{Conclusion}

We developed learning algorithms 
that adapt to the intrinsic dimensionality of the data.
Our algorithms exploit training sets that are close to being low-dimensional
to achieve improved runtime and more optimistic generalization bounds.
For linear classifiers in Euclidean spaces,
we showed data-dependent generalization bounds that can be optimized by PCA,
which is probably the most widely used dimension-reduction technique.
For Lipschitz classifiers in general metric spaces, 
we demonstrated similar data-dependent generalization bounds, 
which suggest a metric analogue of PCA. 
We then designed an algorithm that computes 
the approximate dimensionality of metric data.

An outstanding question left open is to understand the scenarios in which
reducing the dimension (by PCA or other techniques) 
before running a learning algorithm would achieve \emph{better accuracy}
(as opposed to better runtime and generalization bounds).
It would also be interesting to investigate further our notion of 
metric dimension reduction (analogous to PCA), e.g., by devising for it 
better or more practical algorithms, or by finding other contexts where
it could be useful.

\section*{Acknowledgements}
We thank the anonymous referees for constructive suggestions and Ramon van Handel for helpful correspondence.

\bibliographystyle{plain}
%\IfFileExists{alt.bib}{
\bibliography{alt}

\begin{thebibliography}{10}

\bibitem{AK10}
Alexandr Andoni and Robert Krauthgamer.
\newblock The computational hardness of estimating edit distance.
\newblock {\em SIAM J. Comput.}, 39(6):2398--2429, April 2010.

\bibitem{DBLP:journals/ml/BalcanBV06}
Maria-Florina Balcan, Avrim Blum, and Santosh Vempala.
\newblock Kernels as features: On kernels, margins, and low-dimensional
  mappings.
\newblock {\em Mach. Learn.}, 65(1):79--94, 2006.

\bibitem{DBLP:journals/jmlr/BartlettM02}
Peter~L. Bartlett and Shahar Mendelson.
\newblock Rademacher and gaussian complexities: Risk bounds and structural
  results.
\newblock {\em JMLR}, 3:463--482, 2002.

\bibitem{DBLP:journals/jmlr/BiBEBS03}
Jinbo Bi, Kristin~P. Bennett, Mark~J. Embrechts, Curt~M. Breneman, and Minghu
  Song.
\newblock Dimensionality reduction via sparse support vector machines.
\newblock {\em JMLR}, 3:1229--1243, 2003.

\bibitem{bickel07}
Peter~J. Bickel and Bo~Li.
\newblock Local polynomial regression on unknown manifolds.
\newblock In Regina Liu, William Strawderman, and Cun-Hui Zhang, editors, {\em
  Complex Datasets and Inverse Problems}, volume~54 of {\em Lecture
  Notes--Monograph Series}, pages 177--186. Institute of Mathematical
  Statistics, 2007.

\bibitem{blanchard2008}
Gilles Blanchard, Olivier Bousquet, and Pascal Massart.
\newblock Statistical performance of support vector machines.
\newblock {\em The Annals of Statistics}, 36(2):489--531, 2008.

\bibitem{MR2450775}
Gilles Blanchard and Laurent Zwald.
\newblock Finite-dimensional projection for classification and statistical
  learning.
\newblock {\em IEEE Trans. Inform. Theory}, 54(9):4169--4182, 2008.

\bibitem{DBLP:conf/slsfs/Blum05}
Avrim Blum.
\newblock Random projection, margins, kernels, and feature-selection.
\newblock In {\em Subspace, Latent Structure and Feature Selection}, pages
  52--68, 2005.

\bibitem{borg05}
Ingwer Borg and Patrick~J.F. Groenen.
\newblock {\em Modern Multidimensional Scaling: Theory and Applications}.
\newblock Springer, 2005.

\bibitem{DBLP:journals/ftml/Burges10}
Christopher J.~C. Burges.
\newblock Dimension reduction: A guided tour.
\newblock {\em Foundations and Trends in Machine Learning}, 2(4), 2010.

\bibitem{CG06}
R.~Cole and L.-A. Gottlieb.
\newblock Searching dynamic point sets in spaces with bounded doubling
  dimension.
\newblock In {\em 38th annual ACM symposium on Theory of computing}, pages
  574--583, 2006.

\bibitem{der-lee-banach}
Ricky Der and Daniel Lee.
\newblock {Large-Margin Classification in Banach Spaces}.
\newblock In {\em AISTATS 2007}, pages 91--98, 2007.

\bibitem{dudley1999uniform}
Richard~M. Dudley.
\newblock {\em Uniform Central Limit Theorems}.
\newblock Cambridge Studies in Advanced Mathematics. Cambridge University
  Press, 1999.

\bibitem{Enflo69}
P.~Enflo.
\newblock On the nonexistence of uniform homeomorphisms between
  ${L}\sb{p}$-spaces.
\newblock {\em Ark. Mat.}, 8:103--105, 1969.

\bibitem{FBM04}
Kenji Fukumizu, Francis~R. Bach, and Michael~I. Jordan.
\newblock Dimensionality reduction for supervised learning with reproducing
  kernel hilbert spaces.
\newblock {\em JMLR}, 5:73--99, 2004.

\bibitem{semimetric15}
Lee-Ad Gottlieb and Aryeh Kontorovich.
\newblock Nearly optimal classification for semimetrics (arxiv:1502.06208).
\newblock 2015.

\bibitem{DBLP:journals/tit/GottliebKK14}
Lee{-}Ad Gottlieb, Aryeh Kontorovich, and Robert Krauthgamer.
\newblock Efficient classification for metric data.
\newblock {\em {IEEE} Transactions on Information Theory}, 60(9):5750--5759,
  2014.
\newblock Preliminary version in COLT 2010.

\bibitem{DBLP:journals/corr/GottliebK14}
Lee-Ad Gottlieb, Aryeh Kontorovich, and Pinhas Nisnevitch.
\newblock Near-optimal sample compression for nearest neighbors.
\newblock In {\em Neural Information Processing Systems (NIPS)}, 2014.

\bibitem{DBLP:conf/approx/GottliebK10}
Lee-Ad Gottlieb and Robert Krauthgamer.
\newblock Proximity algorithms for nearly-doubling spaces.
\newblock In {\em APPROX-RANDOM}, pages 192--204, 2010.

\bibitem{DBLP:conf/focs/GuptaKL03}
Anupam Gupta, Robert Krauthgamer, and James~R. Lee.
\newblock Bounded geometries, fractals, and low-distortion embeddings.
\newblock In {\em FOCS}, pages 534--543, 2003.

\bibitem{DBLP:journals/jcss/HeinBS05}
Matthias Hein, Olivier Bousquet, and Bernhard Sch{\"o}lkopf.
\newblock Maximal margin classification for metric spaces.
\newblock {\em J. Comput. Syst. Sci.}, 71(3):333--359, 2005.

\bibitem{DBLP:journals/focm/HsuK014}
Daniel Hsu, Sham~M. Kakade, and Tong Zhang.
\newblock Random design analysis of ridge regression.
\newblock {\em Foundations of Computational Mathematics}, 14(3):569--600, 2014.

\bibitem{HA07}
Ke~Huang and Selin Aviyente.
\newblock Large margin dimension reduction for sparse image classification.
\newblock In {\em SSP}, pages 773--777, 2007.

\bibitem{kakade-tewari-lecture}
Sham Kakade and Ambuj Tewari.
\newblock Dudley's theorem, fat shattering dimension, packing numbers.
  {L}ecture 15, {T}oyota {T}echnological {I}nstitute at {C}hicago.
\newblock Available at
  \url{http://ttic.uchicago.edu/~tewari/lectures/lecture15.pdf}, 2008.

\bibitem{MR0124720}
Andre{\u\i}~N. Kolmogorov and Vladimir~M. Tihomirov.
\newblock {$\varepsilon $}-entropy and {$\varepsilon $}-capacity of sets in
  functional space.
\newblock {\em Amer. Math. Soc. Transl. (2)}, 17:277--364, 1961.

\bibitem{MR1892654}
V.~Koltchinskii and D.~Panchenko.
\newblock Empirical margin distributions and bounding the generalization error
  of combined classifiers.
\newblock {\em Ann. Statist.}, 30(1):1--50, 2002.

\bibitem{DBLP:journals/corr/KontorovichW14a}
Aryeh Kontorovich and Roi Weiss.
\newblock A {B}ayes consistent 1-{NN} classifier.
\newblock In {\em AISTATS}, 2015.

\bibitem{NIPS2009_1009}
Samory Kpotufe.
\newblock Fast, smooth and adaptive regression in metric spaces.
\newblock In {\em Advances in Neural Information Processing Systems 22}, pages
  1024--1032, 2009.

\bibitem{DBLP:conf/nips/Kpotufe11}
Samory Kpotufe.
\newblock k-{NN} regression adapts to local intrinsic dimension.
\newblock In {\em Advances in Neural Information Processing Systems 24}, pages
  729--737, 2011.

\bibitem{KpotufeDasgupta2012}
Samory Kpotufe and Sanjoy Dasgupta.
\newblock A tree-based regressor that adapts to intrinsic dimension.
\newblock {\em J. Comput. Syst. Sci.}, 78(5):1496--1515, September 2012.

\bibitem{DBLP:conf/nips/KpotufeG13}
Samory Kpotufe and Vikas~K. Garg.
\newblock Adaptivity to local smoothness and dimension in kernel regression.
\newblock In {\em Advances in Neural Information Processing Systems 26}, pages
  3075--3083, 2013.

\bibitem{KL04}
R.~Krauthgamer and J.~R. Lee.
\newblock Navigating nets: {S}imple algorithms for proximity search.
\newblock In {\em SODA}, pages 791--801, January 2004.

\bibitem{LedouxTal91}
Michel Ledoux and Michel Talagrand.
\newblock {\em Probability in Banach Spaces}.
\newblock Springer-Verlag, 1991.

\bibitem{nonlin-dimred-book}
John~A. Lee and Michel Verleysen.
\newblock {\em Nonlinear Dimensionality Reduction}.
\newblock Information Science and Statistics. Springer, 2007.

\bibitem{DBLP:journals/jmlr/Mendelson03}
Shahar Mendelson.
\newblock On the performance of kernel classes.
\newblock {\em Journal of Machine Learning Research}, 4:759--771, 2003.

\bibitem{MR1965359}
Shahar Mendelson and Roman Vershynin.
\newblock Entropy and the combinatorial dimension.
\newblock {\em Invent. Math.}, 152(1):37--55, 2003.

\bibitem{DBLP:conf/colt/MicchelliP04}
Charles~A. Micchelli and Massimiliano Pontil.
\newblock A function representation for learning in {B}anach spaces.
\newblock In {\em COLT}, pages 255--269, 2004.

\bibitem{mohri-book2012}
Mehryar Mohri, Afshin Rostamizadeh, and Ameet Talwalkar.
\newblock {\em Foundations Of Machine Learning}.
\newblock The MIT Press, 2012.

\bibitem{DBLP:conf/icml/MosciRV07}
Sofia Mosci, Lorenzo Rosasco, and Alessandro Verri.
\newblock Dimensionality reduction and generalization.
\newblock In {\em Machine Learning, Proceedings of the Twenty-Fourth
  International Conference {(ICML} 2007)}, pages 657--664, 2007.

\bibitem{NS07}
Assaf Naor and Gideon Schechtman.
\newblock Planar earthmover is not in $l_1$.
\newblock {\em SIAM J. Comput.}, 37:804--826, June 2007.

\bibitem{DBLP:conf/aistats/PaulBMD13}
Saurabh Paul, Christos Boutsidis, Malik Magdon{-}Ismail, and Petros Drineas.
\newblock Random projections for support vector machines.
\newblock In {\em Proceedings of the Sixteenth International Conference on
  Artificial Intelligence and Statistics, {AISTATS} 2013}, pages 498--506,
  2013.

\bibitem{DBLP:conf/nips/RahimiR07}
Ali Rahimi and Benjamin Recht.
\newblock Random features for large-scale kernel machines.
\newblock In {\em NIPS}, 2007.

\bibitem{DBLP:conf/nips/SabatoST10}
Sivan Sabato, Nathan Srebro, and Naftali Tishby.
\newblock Tight sample complexity of large-margin learning.
\newblock In {\em NIPS}, pages 2038--2046, 2010.

\bibitem{SSSW99}
B.~Sch\"{o}lkopf, J.~{Shawe-Taylor}, A.J. Smola, and R.C. Williamson.
\newblock Kernel-dependent support vector error bounds.
\newblock In {\em ICANN}, 1999.

\bibitem{shwartz2014understanding}
Shai Shalev-Shwartz and Shai Ben-David.
\newblock {\em Understanding Machine Learning: From Theory to Algorithms}.
\newblock Cambridge University Press, 2014.

\bibitem{DBLP:journals/tit/Shawe-TaylorBWA98}
John {Shawe-Taylor}, Peter~L. Bartlett, Robert~C. Williamson, and Martin
  Anthony.
\newblock Structural risk minimization over data-dependent hierarchies.
\newblock {\em IEEE Transactions on Information Theory}, 44(5):1926--1940,
  1998.

\bibitem{DBLP:conf/icml/ShiSHH12}
Qinfeng Shi, Chunhua Shen, Rhys Hill, and Anton van~den Hengel.
\newblock Is margin preserved after random projection?
\newblock In {\em Proceedings of the 29th International Conference on Machine
  Learning, {ICML} 2012}. icml.cc / Omnipress, 2012.

\bibitem{DBLP:journals/tsp/VarshneyW11}
Kush~R. Varshney and Alan~S. Willsky.
\newblock Linear dimensionality reduction for margin-based classification:
  High-dimensional data and sensor networks.
\newblock {\em IEEE Transactions on Signal Processing}, 59(6):2496--2512, 2011.

\bibitem{DBLP:journals/jmlr/LuxburgB04}
Ulrike von Luxburg and Olivier Bousquet.
\newblock Distance-based classification with lipschitz functions.
\newblock {\em Journal of Machine Learning Research}, 5:669--695, 2004.

\bibitem{MR2172729}
Larry Wasserman.
\newblock {\em All of nonparametric statistics}.
\newblock Springer Texts in Statistics. Springer, New York, 2006.

\bibitem{DBLP:journals/tit/WilliamsonSS01}
Robert~C. Williamson, Alex~J. Smola, and Bernhard Sch{\"{o}}lkopf.
\newblock Generalization performance of regularization networks and support
  vector machines via entropy numbers of compact operators.
\newblock {\em {IEEE} Transactions on Information Theory}, 47(6):2516--2532,
  2001.

\bibitem{Y01}
Neal~E. Young.
\newblock Sequential and parallel algorithms for mixed packing and covering.
\newblock In {\em FOCS}, pages 538--546, 2001.

\bibitem{ZXZ09}
Haizhang Zhang, Yuesheng Xu, and Jun Zhang.
\newblock Reproducing kernel {B}anach spaces for machine learning.
\newblock {\em JMLR}, 10:2741--2775, December 2009.

\end{thebibliography}
%}{
%\bibliography{../alt}
%}

\end{document}